\documentclass{article}

\usepackage{PRIMEarxiv}

\usepackage[utf8]{inputenc} % allow utf-8 input
\usepackage[T1]{fontenc}    % use 8-bit T1 fonts
\usepackage{hyperref}       % hyperlinks
\usepackage{url}            % simple URL typesetting
\usepackage{booktabs}       % professional-quality tables
\usepackage{amsfonts}       % blackboard math symbols
\usepackage{nicefrac}       % compact symbols for 1/2, etc.
\usepackage{microtype}      % microtypography
\usepackage{lipsum}
\usepackage{fancyhdr}       % header
\usepackage{graphicx}       % graphics
\graphicspath{{media/}}     % organize your images and other figures under media/ folder
\usepackage{multicol}
\usepackage{amssymb}
\usepackage{graphicx}%
\usepackage{multirow}%
\usepackage{amsmath,amssymb,amsfonts}%
\usepackage{amsthm}%
\usepackage{mathrsfs}%
\usepackage[title]{appendix}%
\usepackage{xcolor}%
\usepackage{textcomp}%
\usepackage{manyfoot}%
\usepackage{booktabs}%
\usepackage[ruled]{algorithm2e}%
\usepackage{algorithmic}
\usepackage{listings}%
\usepackage[justification=centering]{caption}%
\usepackage{makecell}
\usepackage{CJKutf8}
\title{A novel feature selection framework for incomplete data
}

\author{
  Cong Guo,Wei Yang* \\
  Henan University \\
  China \\
  Kaifeng\\
  \texttt{guocong@henu.edu.cn} \\
}

\begin{document}
\maketitle

\begin{abstract}
Feature selection on incomplete datasets is an exceptionally challenging task. Existing methods address this challenge by first employing imputation methods to complete the incomplete data and then conducting feature selection based on the imputed data. Since imputation and feature selection are entirely independent steps, the importance of features cannot be considered during imputation. However, in real-world scenarios or datasets, different features have varying degrees of importance. To address this, we propose a novel incomplete data feature selection framework that considers feature importance. The framework mainly consists of two alternating iterative stages: the M-stage and the W-stage. In the M-stage, missing values are imputed based on a given feature importance vector and multiple initial imputation results. In the W-stage, an improved reliefF algorithm is employed to learn the feature importance vector based on the imputed data. Specifically, the feature importance vector obtained in the current iteration of the W-stage serves as input for the next iteration of the M-stage. Experimental results on both artificially generated and real incomplete datasets demonstrate that the proposed method outperforms other approaches significantly.
\end{abstract}

% keywords can be removed
\keywords{feature selection \and incomplete dataset \and reliefF}

\section{Introduction}
A dataset typically consist of relevant features, irrelevant features, and redundant features. For a given classification or regression task, the presence of irrelevant and redundant features not only increases the model's training time but also diminishes its generalization ability. An effective approach to address this issue is feature selection\cite{bib1}. The goal of feature selection is to sift through the available feature set and identify the subset with the maximum information content, aiming to achieve classification performance close to or higher than that of the original feature set\cite{bib2}. Additionally, feature selection enhances the interpretability of the model.

Feature selection methods can be broadly categorized into three types: filter, wrapper, and embedded methods. Filter methods assign rankings or scores to each feature using various statistical measures. ReliefF\cite{bib3} is the most commonly used filter algorithm\cite{bib4}. Specifically, multiple improved versions of the ReliefF algorithm have been proposed. RReliefF\cite{bib3} is a variant of ReliefF designed for regression problems. RBEFF\cite{bib5} is a ReliefF-based algorithm utilizing random multi-subspaces. $\mu$-relief\cite{bib6} is a mean-based ReliefF algorithm. Since these methods do not depend on any classifier, they exhibit faster computation; however, the feature subsets they select are often not optimal\cite{bib7}.Wrapper methods assess the quality of feature subsets based on the performance of a classifier. They often employ swarm optimization\cite{bib1}, particle swarm optimization\cite{bib8}, whale optimization\cite{bib9}, and other swarm intelligence optimization methods to generate binary solution vectors, where 1 indicates the inclusion of a specific feature and 0 indicates the exclusion. Due to the need for evaluating all possible feature subsets, wrapper methods have high computational costs and are prone to overfitting\cite{bib6}.Embedded methods typically integrate the learning of feature weights into the optimization of the classification model. Additionally, to obtain sparse feature weight vectors, penalty terms related to feature weights are introduced in the loss function. Gradient descent or Alternating Direction Method of Multipliers (ADMM)\cite{bib10} optimization methods are commonly employed to minimize the objective loss. For example, NCFS\cite{bib11} and WKNN-FS\cite{bib12} are embedded feature selection algorithms based on the nearest neighbor model, optimizing the objective function using gradient descent. ERCCFS\cite{bib13} and BLFSE\cite{bib14} utilize the ADMM method to learn feature weight vectors. Compared to wrapper methods, embedded methods often achieve better feature subsets at a smaller computational cost.

The existing feature selection methods are primarily designed to handle complete datasets. However, when missing values are present in the dataset, these methods become inapplicable. In reality, many datasets contain a certain proportion of missing values. According to statistics, 45\% of datasets in the UCI database exhibit missing values\cite{bib15}. For incomplete data, there are feature selection algorithms that can directly address this issue. For instance, in reference\cite{bib16}, a partial distance strategy is first employed to search for neighboring samples, followed by feature selection on incomplete data based on mutual information estimation. Reference\cite{bib17} combines Particle Swarm Optimization (PSO) and the C4.5 classifier to search for the optimal feature subset in incomplete data. However, these methods often fail to achieve satisfactory results.For feature selection on imcomplete data, the prevailing approach is to impute first and then perform feature selection based on the imputed dataset. For example, in reference\cite{bib18}, fuzzy c-means imputation and fuzzy principal component analysis are used to address missing data. Reference\cite{bib19} employs imputation methods such as KNN, EM, mean, along with t-tests and entropy ranking to determine feature weights. Reference\cite{bib20} initially uses KNN, multilayer perceptron, support vector machine for imputing missing data and subsequently employs information gain and genetic algorithms for feature selection. Although these methods yield favorable results, they do not consider the varying importance of different features during the imputation of missing values. Consequently, the imputed data may introduce new biases and even alter the original data distribution\cite{bib21}. To overcome this issue, a viable approach is to perform missing value imputation based on feature importance.

To address this, we propose a novel feature selection framework tailored for incomplete data in this paper. The framework comprises two alternating iterative phases: the Matrix Imputation Stage (M-stage) and the Weight Learning Stage (W-stage). In the M-stage, we introduce a missing value imputation method based on the feature importance vector. This method takes multiple basic imputation results as input to enhance the robustness of the imputation outcomes. It further weights the data features based on the feature importance vector to improve the imputing quality of important feature items. In the W-stage, the $\mu$-reliefA feature selection method is employed to learn the feature importance vector based on the imputed data. Additionally, the feature importance vector output from the current iteration in the W-stage serves as input for the M-stage in the next iteration. The two stages iteratively alternate until the stopping criterion is met.

The remainder of this paper is organized as follows: Section 2 introduces three types of data missing mechanisms and methods for handling missing data. Section 3 provides a detailed description of the proposed algorithm, Section 4 presents experimental results on real-world datasets, and finally, Section 5 concludes this paper.

\section{Preliminaries}
In this section, we first introduce the three missing mechanisms and then provide a brief description of how missing values are handled.

\subsection{Missing mechanisms}
The missingness of data arises from various reasons, and different reasons correspond to different missing mechanisms. In 1976, statisticians Rubin et al. categorized missing mechanisms into three types\cite{bib22}: Missing at Random (MAR), Missing Completely at Random (MCAR), and Missing Not at Random (MNAR). Let the incomplete data set be \({\bf{R}}_{\text{com}} = \{{\bf{R}}_{\text{obs}}, {\bf{R}}_{\text{mis}}\}\), where \({\bf{R}}_{\text{obs}}\) represents the observable data in \({\bf{R}}_{\text{com}}\), and \({\bf{R}}_{\text{mis}}\) represents the missing data in \({\bf{R}}_{\text{com}}\). A binary matrix \({\bf{M}}\) is used to indicate whether data in \({\bf{R}}_{\text{com}}\) are missing. Specifically, \(M_{ij}=0\) indicates the absence of the \(j\)-th feature for the \(i\)-th sample, and otherwise, it indicates non-missing. For Missing at Random (MAR), it implies that \({\bf{M}}\) is independent of the missing values, i.e., \(P({\bf{M}}|{\bf{R}}_{\text{obs}}, {\bf{R}}_{\text{mis}}) = P({\bf{M}}|{\bf{R}}_{\text{obs}})\). For instance, the extent of income missingness varies with the age of respondents, as older or younger respondents are more likely to lack income data. Missing Completely at Random (MCAR) is a special case of MAR, assuming that missingness is independent of the observed values, i.e., \(P({\bf{M}}) = P({\bf{M}}|{\bf{R}}_{\text{com}})\). For example, in RNA sequencing, single-cell RNA sequences exhibit high noise levels, resulting in data being completely randomly missing\cite{bib23}. In a sense, under MCAR, observed data can be considered as a purely random sample from complete data, with similar mean, variance, and overall distribution. For Missing Not at Random (MNAR), it indicates that the missingness depends on unobserved data, and in this case, \(P({\bf{M}}|{\bf{R}}_{\text{com}}) = P({\bf{M}}|{\bf{R}}_{\text{mis}})\). For example, in clinical trials of drugs, patients who voluntarily discontinue medication during the trial are more likely not to record their drug intake\cite{bib24}, leading to MNAR.
\subsection{Handling of missing value}
The presence of missing data can directly or indirectly impact the accuracy of machine learning algorithms. Therefore, handling missing values has consistently held a crucial position in the statistical analysis process. There are primarily three methods for dealing with missing values:

(1){\bf{Deletion Method}}: This involves directly removing samples containing missing values. However, the deletion method can reduce classification performance when there are too many missing values\cite{bib25}.

(2){\bf{Imputation Method}}: This is the most commonly used approach for handling missing values. It involves estimating missing values using a certain strategy and replacing them with the estimated values. Specifically, estimated values can be mean, mode, median, or values obtained based on predictive models\cite{bib26}. Imputation methods can be further subdivided into statistical techniques and machine learning techniques. Statistical imputation methods are predominantly focused on low-rank matrix completion, aiming to find a low-rank matrix that matches the original data matrix at observable data positions. Representative methods include Singular Value Decomposition (SVD)\cite{bib27}, Singular Value Thresholding (SVT)\cite{bib28}, and low-rank matrix completion based on Gaussian Copula (GC-impute)\cite{bib29}. For machine learning-based imputation methods, the simplest and most commonly used is KNN imputation. In recent years, deep learning techniques have introduced imputation methods based on Recurrent Neural Networks (RNN) \cite{bib30,bib31}, Generative Adversarial Networks (GAN)\cite{bib32}, and self-attention\cite{bib33}.

(3){\bf{Ignore Method}}: This method utilizes models capable of directly handling data with missing values for modeling. Therefore, it does not require any preprocessing of missing values. For example, the C4.5 classifier\cite{bib34}, due to incorporating missing situations in the calculation of information gain ratio, can calculate the optimal subset division method for each attribute based on missing data. Additionally, Partial Distance Strategy (PDS)\cite{bib16} can also handle missing data. In this method, when calculating sample similarity, the missing parts are directly ignored, and distance is computed only based on non-missing parts. In summary, because it leverages known information from missing data, the ignore method has advantages over the deletion method.

\section{The proposed method}\label{sec3}
For datasets with missing values, conventional approaches typically involve initially imputing missing values using imputation methods, followed by feature selection based on the imputed dataset. However, existing imputation methods overlook the varying importance of different features during data imputation, potentially introducing biases unfavorable for subsequent feature selection. To address this, we propose a novel incomplete data feature selection framework that considers feature importance. This framework primarily comprises two interrelated stages: the M-stage and the W-stage. Specifically, in the M-stage, missing values are imputed based on given feature importance and multiple imputation results. In the W-stage, an improved reliefF algorithm is employed to learn feature importance based on the imputed data. Notably, the feature importance output from the current iteration in the W-stage serves as input for the M-stage in the next iteration. These two stages iteratively alternate until the stopping criterion for iterations is met. We first provide a brief description of the symbols involved in the algorithm and then offer detailed explanations of the M-stage and W-stage.
\subsection{Notations and definitions}\label{subsec2}
Given a matrix ${\bf{M}}\in{{\bf{R}}^{{n \times m}}}$,we denote its ($i$,$j$) entry,$i$-th row,$j$-th column as ${{\bf{M}}_{ij}}$,${{\bf{M}}_i}$,${{\bf{M}}^j}$ respectively.In this paper,${\bf{X}} = {[{x_1},{x_2}, \cdots ,{x_n}]^T} \in {{\bf{R}}^{n \times m}}$ denotes the matrix of data set with missing values,and ${\bf{y}} = {[{y_1},{y_2}, \cdots ,{y_n}]^T}$ denotes the labels corresponding to the samples.For the feature matrix ${\bf{X}}$, let $\Omega {\rm{ = }}\left\{ {(p,q)\left| {{X_{pq}}{\rm{\ is }}  {\rm{\ observed}}} \right.} \right\}$ be the index set of all observable value elements in ${\bf{X}}$ and $\bar \Omega {\rm{ = }}\left\{ {(p,q)\left| {{X_{pq}} {\rm{\ is \ unobservble}}} \right.} \right\}$ be the index set of all missing value elements in ${\bf{X}}$. $\left\|  \bullet  \right\|_F^2$ denotes the Frobenius paradigm of the matrix. Furthermore, let ${\bf{Z}} \in {{\bf{R}}^{n \times d}}$ be the data matrix of rank $r(r \ll min(n,d))$ obtained after imputing ${\bf{X}}$, which is decomposed into the product of two low-rank matrices ${\bf{Z}} = {\bf{GH}}$, where ${\bf{G}} \in {{\bf{R}}^{n \times r}}$ and ${\bf{H}} \in {{\bf{R}}^{r \times d}}$.

\subsection{M-stage}\label{subsec4}
For incomplete datasets, the goal of the M-stage is to impute missing values in the dataset based on a given feature importance vector ${\bf{v}}$. Specifically, during the initial iteration, we assume ${\bf{v}}$ to be a $d$-dimensional vector with all elements equal to 1, indicating that all features have equal importance. Based on the feature importance vector ${\bf{v}}$, we define the imputation loss function for incomplete data as follows:
\begin{equation}
    \left\| {{\bf{GH}}Diag({\bf{v}}) - {\bf{\hat X}}Diag({\bf{v}})} \right\|_F^2 + \gamma (\left\| {\bf{G}} \right\|_F^2 + \left\| {\bf{H}} \right\|_F^2)
\end{equation}
where ${\rm{Diag}}({\bf{v}}) = \left[ {\begin{array}{*{20}{c}}
v_1&0& \cdots &0\\
0&{{v_2}}& \cdots &0\\
 \vdots & \vdots & \ddots & \vdots \\
0&0& \cdots &{{v_d}}
\end{array}} \right]$,$\gamma>0$ is the regularisation parameter,matrix $\hat {\bf{X}} = {P_\Omega }({\bf{X}}) + {P_{\bar \Omega }}({\bf{GH}})$,the function ${P_\Omega }({\bf{X}})$ is defined as:
\begin{equation}
    {\left[ {{P_\Omega }(X)} \right]_{pq}} = \left\{ \begin{array}{l}
\begin{array}{*{20}{c}}
{{x_{pq}},}&{(p,q) \in \Omega }
\end{array}\\
\begin{array}{*{20}{c}}
{0,}&{{\rm{   }}(p,q) \notin \Omega }
\end{array}
\end{array} \right.
\end{equation}
Note that the product of the two low-rank matrices ${\bf{G}}$ and ${\bf{H}}$ is the populated data matrix ${\bf{Z=GH}}$. Based on the populated matrix ${\bf{Z}}$, the first term $\left\| {{\bf{GH}}Diag({\bf{v}}) - {\bf{\hat X}}Diag({\bf{v}})} \right\|_F^2$ in the loss function can be converted to ${\sum\limits_{(p,q) \in \Omega } {v_q^2\left( {{Z_{pq}} - {X_{pq}}} \right)} ^2}$. This indicates that the padding loss is based on fitting observable data with different feature importance weights.
For missing values in incomplete data, the imputation results typically exhibit significant uncertainty due to the lack of available information. To enhance the robustness of imputation outcomes, we draw inspiration from ensemble learning principles by considering consensus and diversity to determine the missing values. Specifically, let ${\bf{X}}^{(1)},...,{\bf{X}}^{(m)}$ denote the imputed results obtained from m different imputation methods. In order to fit these results, the following loss term can be introduced:
\begin{equation}
    \frac{1}{m}\sum\limits_{i = 1}^m {\left\| {{\bf{Z}} - {{\bf{X}}^{(i)}}} \right\|_F^2} 
\end{equation}
By combining Eqs.(1) and (3), we can formalise the imputation of incomplete data as the following optimisation problem:
\begin{equation}
    \mathop {\min }\limits_{{\bf{G}},{\bf{H}}} \frac{1}{m}\sum\limits_{i = 1}^m {\left\| {{\bf{GH}} - {{\bf{X}}^{(i)}}} \right\|_F^2}  + \left\| {{\bf{GH}}Diag({\bf{v}}) - {\bf{\hat X}}Diag({\bf{v}})} \right\|_F^2 + \gamma (\left\| {\bf{G}} \right\|_F^2 + \left\| {\bf{H}} \right\|_F^2)
\end{equation}
For the optimization problem above, we employ an alternating iterative approach for solution. Specifically, we initially initialize {\bf{G}} as a randomly generated matrix \({\bf{G}}^{(0)}\) with orthogonal columns. Subsequently, in the $k$-th iteration, we first compute ${\bf{H}}^{(k)}$ based on ${\bf{G}} = {\bf{G}}^{(k-1)}$, and then compute ${\bf{G}}^{(k)}$ based on ${\bf{H}} = {\bf{H}}^{(k)}$, repeating this process until the stopping criterion is met.

When ${\bf{G}}$ is held fixed at ${\bf{G}}^{(k-1)}$, the optimization problem in Equation (4) can be reformulated as follows:
\begin{equation}
    {{\bf{H}}^{(k)}} = \mathop {\arg \min }\limits_{{\bf{H}} \in {{\bf{R}}^{r \times d}}} \frac{1}{m}\sum\limits_{i = 1}^m {\sum\limits_{q = 1}^d {\left\| {{{\bf{G}}^{(k - 1)}}{{\bf{H}}^q} - {{\left( {{{\bf{X}}^{(i)}}} \right)}^q}} \right\|_F^2} }  + \sum\limits_{q = 1}^d {v_q^2\left\| {{{\bf{G}}^{(k - 1)}}{{\bf{H}}^q} - {{\hat {\bf{X}}}^q}} \right\|_2^2 + } \gamma \left\| {{{\bf{H}}^q}} \right\|_2^2
\end{equation}
And Eq.(5) can be decomposed into $d$ independent optimisation subproblems:
\begin{equation}
    \mathop {\arg \min }\limits_{{{\bf{H}}^q} \in {{\bf{R}}^{r \times 1}}} \frac{1}{m}\sum\limits_{i = 1}^m {\left\| {{{\bf{G}}^{(k - 1)}}{{\bf{H}}^q} - {{\left( {{{\bf{X}}^{(i)}}} \right)}^q}} \right\|_2^2}  + v_q^2\left\| {{{\bf{G}}^{(k - 1)}}{{\bf{H}}^q} - {{\hat {\bf{X}}}^q}} \right\|_2^2 + \gamma \left\| {{{\bf{H}}^q}} \right\|_2^2{\rm{      }},q = 1,2, \ldots ,d
\end{equation}
For the $q$-th subproblem, making its derivative with respect to the parameter ${\bf{H}}^q$ equal to 0 yields the analytical solution.
\begin{equation}
    {\left( {{{\bf{H}}^{(k)}}} \right)^q} = {\left[ {\left( {v_q^2 + 1} \right){{\left( {{{\bf{G}}^{(k - 1)}}} \right)}^T}{{\bf{G}}^{(k - 1)}} + \gamma {{\bf{I}}_r}} \right]^{ - 1}}\left[ {v_q^2{{({{\bf{G}}^{(k - 1)}})}^T}{{\hat {\bf{X}}}^q} + \frac{1}{m}{{\left( {{{\bf{G}}^{(k - 1)}}} \right)}^T}\sum\limits_{i = 1}^m {{{\left( {{{\bf{X}}^{(i)}}} \right)}^q}} } \right]
\end{equation}
where ${\bf{I}}_r$ is a unit matrix of size $r×r$. When ${\bf{H}}$ is fixed to ${\bf{H}}^{(k)}$, the optimisation problem in Eq.(7) can be converted to:
\begin{equation}
{{\bf{G}}^{(k)}} = \mathop {\arg \min }\limits_{{{\bf{G}}_p} \in {{\bf{R}}^{1 \times r}}} \frac{1}{m}\sum\limits_{i = 1}^m {\sum\limits_{p = 1}^n {\left\| {{{\bf{G}}_p}{{\bf{H}}^{(k)}} - {{\left( {{{\bf{X}}^{(i)}}} \right)}_p}} \right\|} _2^2}  + \left\| {{{\bf{G}}_p}{{\bf{H}}^{(k)}}{\mathop{\rm Diag}\nolimits} ({\bf{v}}) - {{\hat {\bf{X}}}_p}{\mathop{\rm Diag}\nolimits} ({\bf{v}})} \right\|_2^2 + \gamma \left\| {{{\bf{G}}}} \right\|_2^2
\end{equation}
Similarly, Eq.(8) can be decomposed into $n$ independent optimisation subproblems:
\begin{equation}
    \mathop {\arg \min }\limits_{{{\bf{G}}_p} \in {{\bf{R}}^{1 \times r}}} \frac{1}{m}\sum\limits_{i = 1}^m {\left\| {{{\bf{G}}_p}{{\bf{H}}^{(k)}} - {{\left( {{{\bf{X}}^{(i)}}} \right)}_p}} \right\|_2^2}  + \left\| {{{\bf{G}}_p}{{\bf{H}}^{(k)}}{\mathop{\rm Diag}\nolimits} ({\bf{v}}) - {{\hat {\bf{X}}}_p}{\mathop{\rm Diag}\nolimits} ({\bf{v}})} \right\|_2^2 + \gamma \left\| {{{\bf{G}}_p}} \right\|_2^2{\rm{      }},p = 1,2,...,n
\end{equation}

For the $p$-th subproblem, making its derivative with respect to the parameter ${\bf{G}}_p$ 0 yields an analytical solution:
\begin{equation}
    {\left( {{{\bf{G}}^{(k)}}} \right)_p} = \left[ {{{\hat {\bf{X}}}_p}{\mathop{\rm Diag}\nolimits} ({\bf{v}}){{({{\bf{H}}^{(k)}}{\mathop{\rm Diag}\nolimits} ({\bf{v}}))}^T} + \frac{1}{m}\sum\limits_{i = 1}^m {{{\left( {{{\bf{X}}^{(i)}}} \right)}_p}{{({{\bf{H}}^{(k)}})}^T}} } \right]{\left[ {{{\bf{H}}^{(k)}}{\mathop{\rm Diag}\nolimits} ({\bf{v}}){{({{\bf{H}}^{(k)}}{\mathop{\rm Diag}\nolimits} ({\bf{v}}))}^T} + \gamma {{\bf{I}}_r} + {{\bf{H}}^{(k)}}{{({{\bf{H}}^{(k)}})}^T}} \right]^{ - 1}}
\end{equation}

In this paper, we set the convergence condition for the M-stage:
\begin{equation}
    \left| {ob{j_{(k)}} - ob{j_{(k - 1)}}} \right| < \eta 
\end{equation}
where $obj_k$ is the value of the objective function of Eq.(4) after the $k$-th iteration of the M-stage and the threshold $\eta$ is a smaller positive number. When this stage converges, the populated data matrix can be obtained based on ${\bf{Z}} = {\bf{G}}^{(k)}{\bf{H}}^{(k)}$. In particular, we call this method the integration-based weighted matrix filling method (EWMC).

\subsection{W-stage}\label{subsec5}
The objective of the W-stage is to learn the feature importance vector based on the imputed feature matrix ${\bf{Z}}$ and the corresponding label vector. Consequently, any supervised feature selection algorithm capable of obtaining a feature importance vector can be employed in this stage. Considering the successful applications of ReliefF-type feature selection algorithms in various domains, we opt for its latest variant, $\mu$-relief\cite{bib6}, to learn the feature importance vector in this study.

$\mu$-relief is a feature selection algorithm based on average distances. For a given dataset, $\mu$-relief initializes the feature importance vector ${\bf{v}}$ as a vector of all ones. In each iteration, it updates the vector based on a randomly chosen individual sample $(x_i, y_i)$. Specifically, the set comprised of other samples in the dataset with the same label as $x_i$ is defined as the hit set ($H_i$).For the set $M_{ic}$ consisting of samples in the dataset with category label $c({c \ne {y_i}})$ is called the missing set. For given two samples $x_i$ and $x_j$, we denote the Euclidean distance between them by $d_{ij}$. Thus, the average of the distances between sample $x_i$ and all the samples in its hit set $H_i$ and in each miss set $M_{ic}$ can be expressed respectively as ${d_{{H_i}}} = \frac{1}{{\left| {{H_i}} \right|}}\sum\limits_{{x_j} \in {H_i}} {{d_{ij}}} $ and ${d_{{M_{ic}}}} = \frac{1}{{\left| {{M_{ic}}} \right|}}\sum\limits_{{x_j} \in {M_{ic}}} {{d_{ij}}} $,where $\left| {{H_i}} \right|$ and $\left| {{M_{ic}}} \right|$ denote the number of samples in the sets $H_i$ and $M_{ic}$, respectively. For the $l$-th feature $f_l$, we update the weights $v_l$ according to the following equation:
\begin{equation}
    {v_l} = {v_l} - \frac{1}{{\left| {{H_i}} \right|}}\sum\limits_{{x_j} \in {H_i}} {diff({x_i},{x_j},{f_l})} \left| {{d_{ij}} - {d_{{H_i}}}} \right| + \sum\limits_{c \ne {y_i}} {\frac{1}{{n({y_c})}}} \sum\limits_{{x_j} \in {M_{ic}}} {diff({x_i},{x_j},{f_l})} \left| {{d_{ij}} - {d_{{M_{ic}}}}} \right|
\end{equation}
where $n(y_j)$ is the number of all samples in the data set with category $y_j$. The $diff$ function is defined as follows:
\begin{equation}
    diff({x_i},{x_j},{f_l}) = {\rm{ }}\left\{ {\begin{array}{*{20}{c}}
{\frac{{\left| {{x_i}[l] - {x_j}[l]} \right|}}{{\max ({f_l}) - \min ({f_l})}},if{\rm{ }}{f_l}{\rm{\ is\ continuous\,}}}\\
{{\rm{0, }}if{\rm{ }}{f_l}{\rm{\ is \ discrete \ and \ }}{x_i}[l] = {x_j}[l],}\\
{{\rm{1,}}if{\rm{ }}{f_l}{\rm{\  is \ discrete \ and\ }}{x_i}[l] \ne {x_j}[l],}
\end{array}} \right.{\rm{ }}
\end{equation}
It should be noted that the update procedure for the weight vector in this paper differs from the original $\mu$-relief algorithm. We refer to the modified algorithm as $\mu$-reliefA. The key modifications include two aspects: (1) the introduction of the absolute difference of distances in the weight update; (2) consideration of all samples during the weight update. Algorithm 1 provides the pseudocode for the $\mu$-reliefA algorithm. Additionally, Algorithm 2 presents the pseudocode for the incomplete data feature selection framework.
\begin{algorithm}
\LinesNumbered

\caption{$\mu$-reliefA for feature selection}%\label{alg:two}
\KwIn{${\bf{X}} \in {{\bf{R}}^{n \times d}}$:feature matrix, ${\bf{y}} \in {{\bf{R}}^{n \times 1}}$:label vector;}
\KwOut{feature importance vector $\bf{v}$}
Initialize $\textbf{v}$=$(1,...,1)$;\\
\For{$i\leftarrow 1$ \KwTo $m$}{
Randomly select a sample $x_i$;\\
Construct Hit set($H_{i}$) and Miss sets($M_{ic}$);\\
Compute ${D_{H_{i}}}$ and ${D_{{M_{ic}}}}$;\\
\For{$l\leftarrow 1$ \KwTo $d$}{
Update $v_l$ according to Eq(12);
}

}
{return $\bf{v}$}
%\Comment{aaa}
\end{algorithm}

\begin{algorithm}
\LinesNumbered
\caption{The proposed framework for feature selection on incomplete datasets}%\label{alg:two}

\KwIn{$\textbf{X}$($n\times d$): feature matrix with missing values, $\textbf{y}$($n\times 1$): label vector, ${\bf{X}}^{(1)},...,{\bf{X}}^{(m)}:m$ complete data matrix imputed by other methods, $r$: rank of $\textbf{G}$ and $\textbf{H}$,   $\gamma$: regularization parameter, $\Delta$: small positive constant, $\eta$: convergence threshold for $M$-stage;}
Initialization:{$\zeta^{(0)}=-\infty$},{$t=0$,$\textbf{v}=\{1,...,1\}$;}\\
\Repeat{$|\zeta^{(t)}-\zeta^{(t-1)}|<\Delta$}{
{\bf{M-stage}}:\\
Initialize $ \textbf{G}^{(0)} $as a random matrix;\\
$k=0$;\\
\Repeat{the conditions in Eq.(13) are satisfied}{
$k=k+1$;\\
\For{$q\leftarrow 1$ \KwTo $d$}{
Compute ${({{\bf{H}}^{(k)}})^q}$according to $\textbf{v}^{(t)}$and Eq.(7);\\
}
\For{$p\leftarrow 1$ \KwTo $n$}{
Compute ${({{\bf{G}}^{(k)}})_p}$ according to $\textbf{v}^{(t)}$ and Eq.(10);\\
}
{${\textbf{Z}}= {{\bf{G}}^{(k)}}{{\bf{H}}^{(k)}} $};\\
}
{\bf{W-stage}}:\\
$t=t+1$;\\
Use $\mu$-reliefA to compute $\textbf{v}^{(t)}$ based on ${\textbf{Z}}$ and $\textbf{y}$;\\
{$\zeta^{(t)}= \left\| {{{\bf{\textbf{v}}}^{(t)}}} \right\|_2^2$};
}
{return $\textbf{v}^{(t)}$}

%\Comment{aaa}
\end{algorithm}

\begin{table*}
\caption{Datasets}
    \label{tab:my_label}
    \centering
    \renewcommand\arraystretch{1.5}
    \begin{tabular}{cccccc}
    \hline
        Number&	Dataset&	Instances&	Features&	Classes&	Missing rate\\
        \hline
1&	Wine&	178	&13	&3	&0\\
2&	Spect&	267&	22	&2&	0\\
3&	heart&	303&	13&	2&	0\\
4&	Thoracic Surgery&	470&	16&	2&	0\\ \hline
5&Autistic Spectrum&	292&	20&	2&	1.54\\
6&	HCC survival& 	165&	49&	2&	10.21\\
7&	heart-h&	294&	12&	2&	13.91\\
8&	pbc&	418&	18&	2&	16.46\\
9&	horse&	300&	27&	2&	19.81\\
10&	colic&	368&	22&	2&	23.8  \\
        \hline
    \end{tabular}
    
\end{table*}

\section{Experiment}\label{sec4}

In this section, we first introduce the dataset used for the experiment, then describe the experimental setup, and finally give the results.
\subsection{Datasets}\label{sec1}
To assess the performance of the proposed method, we utilized 10 real datasets from UCI\cite{bib35}. Table 1 provides detailed information about the datasets, where the first four datasets are complete, and the remaining six datasets contain missing values to varying degrees. To transform the first four datasets into incomplete datasets, missing values were generated based on three missing rates (5\%, 10\%, 15\%).

\subsection{Experiment settings}
For the proposed framework, in the M-stage, it is necessary to obtain multiple imputation results for missing data based on various imputation methods. In the experiments, we employed three imputation methods: Expectation-Maximization (EM) imputation, fast k-Nearest Neighbors ($k$=5) imputation, and Singular Value Decomposition (SVD)\cite{bib27}. The convergence parameter $\eta$ was set to 0.1, and parameters $r$ and $\gamma$ were set to 5 and 20, respectively. For each dataset, we conducted a five-fold cross-validation experiment. Specifically, for a given training and validation split, the proposed method first imputed missing values for the training data, then learned the feature importance vector based on the imputed training data, and finally imputed missing values for the test set based on this feature importance vector. On the imputed test set, we calculated the classification accuracy of the k-Nearest Neighbors ($k$=5) classifier. The final results represent the averages of 10 runs of the five-fold cross-validation experiments.

\subsection{Experimental Results}\label{sec2}
The proposed method in the M-stage utilizes EWMC for imputing missing values in incomplete data, while in the W-stage, it employs the $\mu$-reliefA algorithm to learn the feature importance vector based on the imputed dataset. To assess the proposed method, we compare it with seven combination methods: Fast KNN+$\mu$-reliefA, SVD+$\mu$-reliefA, EM+$\mu$-reliefA, Sinkhorn+$\mu$-reliefA, EWMC+$\mu$-relief, EWMC+reliefF, and EWMC+NCFS. Specifically, fast KNN is a fast k-Nearest Neighbors-based imputation method, SVD is a Singular Value Decomposition-based imputation method, EM is an Expectation-Maximization imputation method based on maximum likelihood estimation, Sinkhorn is an imputation method that uses end-to-end learning to minimize optimal transport loss\cite{bib36}. Moreover, $\mu$-relief\cite{bib6},reliefF\cite{bib3},and NCFS\cite{bib11} are three feature selection methods based on nearest neighbor models.

In this section, we first conduct experimental comparisons on both artificial and real missing data sets. Subsequently, we investigate the parameter sensitivity of the proposed algorithm and finally explore the convergence of the EWMC method.

\subsubsection{Comparison experiments on the datasets with artificially generated missing values.}
To evaluate the performance of the proposed method, we conducted comparative experiments on datasets with artificially generated missing values. For the datasets Wine, Spect, Heart, and Thoracic Surgery, we constructed six missing data scenarios based on two missing mechanisms (completely random missing and non-random missing) and three missing rates (5\%, 10\%, 15\%). Figs.1 and 2 illustrate the classification accuracy of eight methods on datasets constructed under completely random and non-random missing mechanisms with varying numbers of features. From the figures, it can be observed that the proposed method, EWMC+$\mu$-reliefA, generally achieved better results in most scenarios. For convenience of comparison, we further calculated the average accuracy under different numbers of features for datasets constructed under both missing mechanisms. The statistical results for the two missing mechanisms are presented in Tables 2 and 3. It can be seen from the tables that the proposed method obtained the best results 12 and 10 times in Tables 2 and 3, respectively, indicating its significant superiority over the other seven methods. Additionally, we observed that on the heart dataset with MNAR data, as the missing rate increased, the average classification accuracy of almost all methods we used increased, contrary to previous expectations. We speculate that this phenomenon may be attributed to the reduction of noise values on the heart dataset due to non-random missing, thereby increasing the predictive classification accuracy.
\begin{figure}[!]
  \centering
  \includegraphics[width=\linewidth]{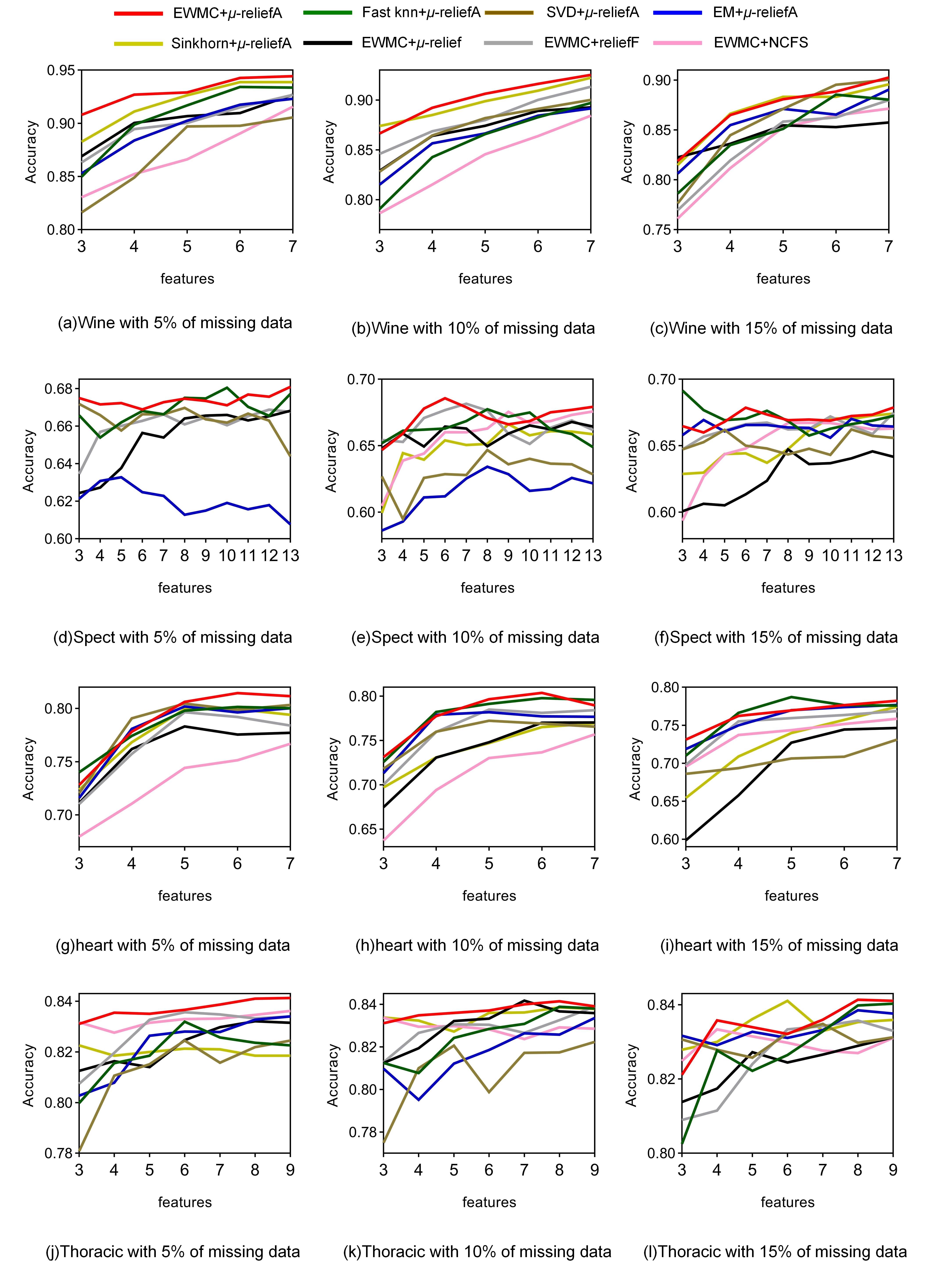}
  \caption{Comparison results of eight methods on four datasets under MCAR.}\centering
\end{figure}

\begin{figure}[!]
  \centering
  \includegraphics[width=\linewidth]{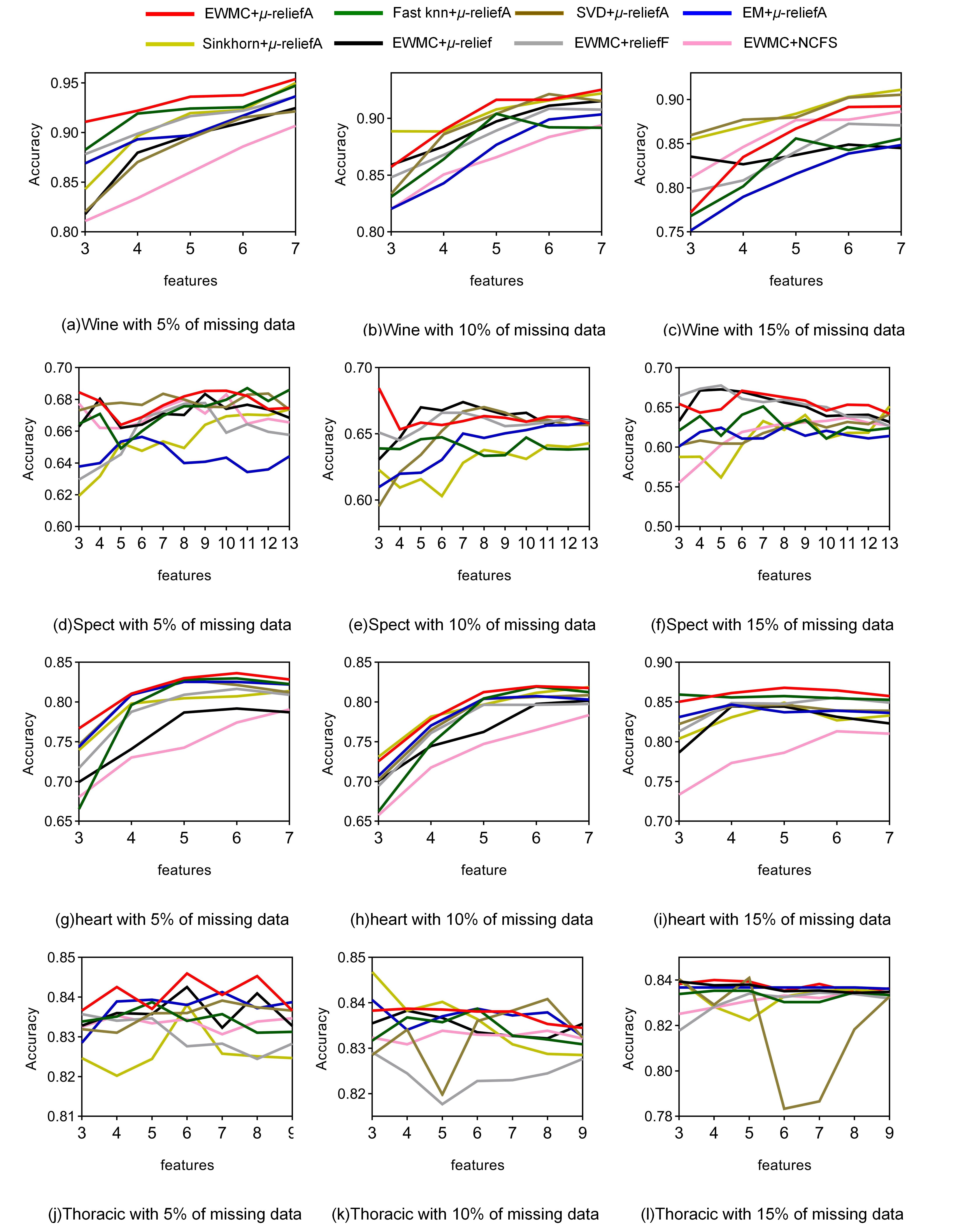}
  \caption{Comparison results of eight methods on four datasets under MNAR.}\centering
\end{figure}
\subsubsection{Comparison experiments on the datasets with real missing values.}
In this section, we investigated the performance of the proposed method on six real-world datasets with missing values: Autistic Spectrum, HCC survival, heart-h, pbc, horse, and colic. Fig.3 presents the experimental results of the proposed method and seven other algorithms on these six datasets. From the figure, it can be observed that the proposed method achieved the best results only on the horse dataset. Although EWMC + $\mu$-relief performed similarly to the proposed method on the heart-h and pdc datasets, the classification accuracy of EWMC+$\mu$-relief was consistently inferior to the proposed method on the other four datasets: Autistic Spectrum,HCC survival,horse,and colic. This indicates the effectiveness of our improvement to $\mu$-relief. It is also noteworthy that the performance of EWMC + NCFS and Sinkhorn+$\mu$-reliefA was highly unstable. Furthermore, Table 4 provides the average results for the six datasets under different numbers of features. The proposed algorithm obtained the best results four times in Table 4, including one in terms of averages.

\begin{table*}
\renewcommand\arraystretch{1.5}
  \caption{Average ACC of the datasets with MCAR data.}\centering
  \label{tab:commands}
  \begin{tabular*}{1.03\linewidth}{l|l|llllllll}%{ccl}
  \hline

{\small Data} &
  {\small Missing} &
{{\small EWMC+}}  &
  {\small Fast knn+} &
{\small SVD+} &
 {\small EM+} &
 {\small EWMC+} &
 {\small EWMC+} &
 {\small EWMC+} &
 {\small Sinkhorn+} 
  \\ 
                         {\small sets}   & {\small rate}     & $\mu$-reliefA        &$\mu$-reliefA  &$\mu$-reliefA  &$\mu$-reliefA &$\mu$-relief&reliefF&NCFS &$\mu$-reliefA                      \\
    %\thead{Data \\ sets}  &\thead{Missing \\ rate}& IWMC &\\
 
    \hline
                   \multirow{3}*{\rotatebox{90}{\small Wine} } & \makecell[c]{5\%}   & \small0.93	&\small0.906	&\small0.873	&\small0.895	&\small0.902&\small	0.899&\small0.87&\small0.919\\
                   
    & \makecell[c]{10\%}   & \small0.901& \small	0.855& \small	0.873	& \small0.862& \small	0.869& \small	0.881& \small	0.839& \small	0.897\\
                    & \makecell[c]{15\%}  & \small0.871& \small	0.847& \small	0.857& \small	0.857&\small	0.844& \small	0.837& \small	0.832& \small	0.868\\
\cmidrule{1-2}
           \multirow{3}*{\rotatebox{90}{\small Spect} }         & \makecell[c]{5\%}   & \small0.673&\small	0.669&\small	0.663&\small	0.62&\small	0.653&\small	0.66&\small	0.66&\small	0.66
\\
    & \makecell[c]{10\%}   &\small0.671&\small	0.663&\small	0.629&\small	0.615&\small0.659&\small	0.665&\small	0.657&\small	0.649
\\
                    & \makecell[c]{15\%}  &\small0.67&\small	0.67&\small	0.651&\small	0.663&\small	0.627&\small	0.663&\small	0.651&\small	0.652
\\
\cmidrule{1-2}
\multirow{3}*{\rotatebox{90}{\small \small heart} }& \makecell[c]{5\%}   &\small0.787&\small	0.782&\small	0.783&\small	0.779&\small	0.761&\small	0.768&\small	0.73&\small	0.777
\\
    & \makecell[c]{10\%}   & \small0.779& \small	0.778& \small	0.756& \small	0.765& \small0.738& \small	0.762& \small	0.71& \small	0.741
\\
                    & \makecell[c]{15\%}  & \small0.764& \small	0.763& \small	0.705& \small	0.757& \small	0.694& \small	0.748& \small	0.737& \small	0.726
\\
\cmidrule{1-2}
\multirow{3}*{\rotatebox{90}{\small \small Thoracic} }& \makecell[c]{5\%}   & \small0.837& \small	0.819& \small	0.813& \small	0.822& \small	0.828& \small	0.823& \small	0.832& \small	0.82
\\
    & \makecell[c]{10\%}   & \small0.837& \small	0.825& \small0.808& \small	0.817& \small	0.83& \small	0.828& \small	0.828& \small	0.834\\
                    & \makecell[c]{15\%}  & \small0.834& \small	0.827& \small	0.83& \small	0.833& \small	0.825& \small	0.824& \small	0.829& \small	0.834

\\
\hline
& \makecell[c]{Win/Tie}  & \small	9/3& \small	0/2& \small	0/0& \small	0/0& \small	0/1& \small	0/0& \small	0/0& \small	0/0		\\
    \hline
  \end{tabular*}
\end{table*}

\begin{table*}
\renewcommand\arraystretch{1.5}
  \caption{Average ACC of the datasets with MNAR data.}\centering
  \label{tab:commands}
  \begin{tabular*}{1.03\linewidth}{l|l|llllllll}%{ccl}
  \hline

{\small Data} &
  {\small Missing} &
{{\small EWMC+}}  &
  {\small Fast knn+} &
{\small SVD+} &
 {\small EM+} &
 {\small EWMC+} &
 {\small EWMC+} &
 {\small EWMC+} &
 {\small Sinkhorn+} 
  \\ 
                         {\small sets}   & {\small rate}     & $\mu$-reliefA       &$\mu$-reliefA  &$\mu$-reliefA  &$\mu$-reliefA &$\mu$-relief&reliefF&NCFS &$\mu$-reliefA                       \\
    %\thead{Data \\ sets}  &\thead{Missing \\ rate}& IWMC &\\
 
    \hline
                   \multirow{3}*{\rotatebox{90}{\small Wine} } & \makecell[c]{5\%}   & \small0.932& \small	0.919& \small	0.884& \small	0.902& \small	0.886& \small	0.91& \small	0.859& \small	0.906
\\
                   
    & \makecell[c]{10\%}   & \small0.901& \small	0.876& \small	0.892& \small	0.868& \small	0.891& \small	0.884& \small	0.862& \small	0.904
\\
                    & \makecell[c]{15\%}  & \small0.851& \small	0.824& \small	0.884& \small0.808& \small	0.838& \small	0.837& \small	0.859& \small	0.884\\
\cmidrule{1-2}
           \multirow{3}*{\rotatebox{90}{\small Spect} }         & \makecell[c]{5\%}   & \small0.673&\small	0.669&\small	0.663&\small0.62&\small	0.653&\small	0.66&\small0.66&\small	0.66

\\
    & \makecell[c]{10\%}   &\small0.661&\small	0.66&\small0.64&\small	0.657&\small	0.641&\small	0.649&\small	0.64&\small	0.628\\
                    & \makecell[c]{15\%}  &\small0.654&\small	0.627&\small0.615&\small	0.654&\small	0.62&\small	0.651&\small	0.615&\small	0.612

\\
\cmidrule{1-2}
\multirow{3}*{\rotatebox{90}{\small \small heart} }& \makecell[c]{5\%}   &\small0.79&\small	0.769&\small	0.777&\small	0.778&\small	0.761&\small	0.769&\small	0.734&\small	0.787

\\
    & \makecell[c]{10\%}   & \small0.814& \small	0.788& \small	0.803& \small	0.805& \small	0.761& \small	0.787& \small	0.743& \small0.792

\\
                    & \makecell[c]{15\%}  & \small0.86& \small	0.856& \small0.838& \small	0.838& \small	0.825& \small	0.842& \small	0.783& \small	0.828

\\
\cmidrule{1-2}
\multirow{3}*{\rotatebox{90}{\small \small Thoracic} }& \makecell[c]{5\%}   & \small0.84& \small	0.834& \small	0.835& \small	0.837& \small	0.836& \small	0.83& \small	0.833& \small	0.826

\\
    & \makecell[c]{10\%}   & \small0.837& \small	0.834& \small0.832& \small	0.836& \small	0.834& \small	0.824& \small	0.832& \small	0.835
\\
                    & \makecell[c]{15\%}  & \small0.837& \small	0.833& \small	0.818& \small0.836& \small	0.836& \small	0.83& \small	0.831& \small	0.832

\\
\hline
& \makecell[c]{Win/Tie}  & \small9/1& \small	0/0& \small	0/0& \small	0/1& \small	2/0& \small	0/0& \small	0/0& \small	0/0	\\
    \hline
  \end{tabular*}
\end{table*}

\begin{figure*}[!]
  \centering
  \includegraphics[width=\linewidth]{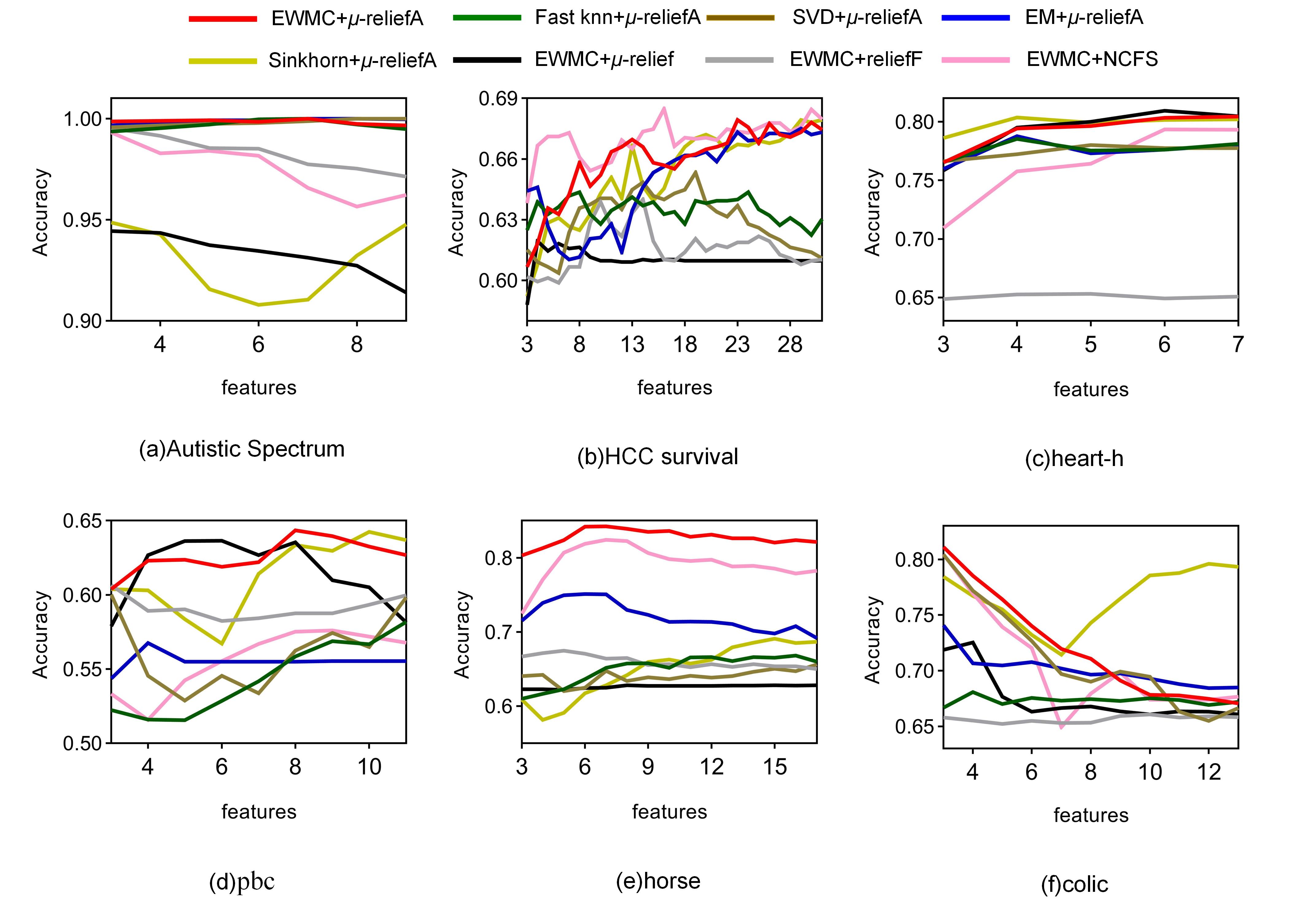}
  \caption{Comparison results of eight methods on six datasets with real missing values.}\centering
\end{figure*}

\begin{table*}
\renewcommand\arraystretch{1.5}
  \caption{Average ACC of the datasets with real missing data.}\centering
  \label{tab:commands}
  \begin{tabular*}{1.03\linewidth}{l|llllllll}%{ccl}
  \hline

{\small Data} &

{{\small EWMC+}}  &
  {\small Fast knn+} &
{\small SVD+} &
 {\small EM+} &
 {\small EWMC+} &
 {\small EWMC+} &
 {\small EWMC+} &
 {\small Sinkhorn+} 
  \\ 
                         {\small sets}       & $\mu$-reliefA        &$\mu$-reliefA  &$\mu$-reliefA  &$\mu$-reliefA &$\mu$-relief&reliefF&NCFS &$\mu$-reliefA                                         \\
    %\thead{Data \\ sets}  &\thead{Missing \\ rate}& IWMC &\\
 
    \hline
                  {\small Autistic Spectrum}  & \small0.998& \small	0.996& \small	0.998& \small	0.998& \small	0.933& \small0.983& \small	0.975& \small0.929
\\
  
\cmidrule{1-1}

{\small HCC survival}  & \small0.659& \small	0.634& \small	0.629& \small	0.615& \small0.649& \small	0.61& \small	0.67& \small	0.652

\\
\cmidrule{1-1}
{\small heart-h}  & \small0.792& \small	0.776& \small0.774& \small	0.775& \small	0.793& \small	0.65& \small	0.763& \small	0.798

\\
\cmidrule{1-1}
{\small pbc}  & \small0.626& \small	0.544& \small	0.561& \small	0.555& \small	0.615& \small	0.591& \small	0.556& \small	0.612

\\
\cmidrule{1-1}
{\small horse}  & \small0.827& \small	0.65& \small0.64& \small	0.66& \small0.72& \small	0.626& \small	0.792& \small0.649

\\
\cmidrule{1-1}
{\small colic}  & \small0.72& \small	0.672& \small	0.71& \small	0.7& \small0.675& \small	0.656& \small	0.705& \small	0.765

\\

\hline
{Win/Tie}  & \small3/1& \small	0/0& \small	0/1& \small	0/1& \small	2/0& \small	0/0& \small	0/0& \small	0/0	\\
    \hline
  \end{tabular*}
\end{table*}

\subsubsection{Wilcoxon signed-rank test results}
Based on the results from Figs.1-3, we employed the Wilcoxon signed-rank test\cite{bib37} to assess the statistical significance differences between the proposed algorithm and the comparison algorithms across a total of 243 instances. Table 5 presents the statistical results. The Wilcoxon signed-rank test is a non-parametric hypothesis test for comparing two paired groups. When the p-value is greater than 0.05, it is considered that there is no significant difference between the two methods; otherwise, it indicates a significant difference between the two groups of data. In this study, NO.R+ represents the number of instances where EWMC+$\mu$-reliefA outperformed the comparison method, and NO.R- represents the number of instances where EWMC+$\mu$-reliefA performed worse than the comparison method. From the statistical results in Table 5, it is evident that all NO.R+ values are significantly greater than NO.R-, and all p-values are less than 0.05. This indicates that the proposed method significantly outperforms the other comparison methods.

\begin{table*}
\renewcommand\arraystretch{1.5}
  \caption{The Wilcoxon signed-rank test results on classification accuracy. (‘yes’ means the significant difference between the performance of two methods).}\centering
  \label{tab:commands}
  \begin{tabular*}{0.85\linewidth}{lllll}%{ccl}
  \hline

{\small Pairwise comparison} &
{\small NO.R+ }  &
  {\small NO.R-} &
  {\small P-Value} &
  {\small Significant difference} 
  \\ 
                         
    %\thead{Data \\ sets}  &\thead{Missing \\ rate}& IWMC &\\
 
    \hline
                  {\small EWMC+$\mu$-reliefA VS. Fast knn+$\mu$-reliefA}  & \small207& \small	36& \small	0.000177& \small	yes\\
{\small EWMC+$\mu$-reliefA VS. SVD+$\mu$-reliefA}  & \small223& \small	20& \small	9.43853e-07& \small	yes\\
{\small EWMC+$\mu$-reliefA VS. EM+$\mu$-reliefA}  & \small215& \small	28& \small	3.96374e-06& \small	yes\\
{\small EWMC+$\mu$-reliefA VS. EWMC+$\mu$-relief}  & \small217& \small	26& \small	0.0001103& \small	yes\\
{\small EWMC+$\mu$-reliefA VS. EWMC+reliefF}  & \small230& \small	13& \small	1.125986e-07& \small	yes\\
{\small EWMC+$\mu$-reliefA VS. EWMC+NCFS}  & \small209& \small	34& \small	0.0019947& \small	yes\\
{\small EWMC+$\mu$-reliefA VS. Sinkhorn+$\mu$-reliefA}  & \small200& \small	43& \small	0.0005269& \small	yes\\

\hline

  \end{tabular*}
\end{table*}

\subsubsection{Parameter sensitivity analysis}
Based on the heart and Wine datasets, we investigated the influence of parameters $r$ and $\gamma$ on the performance of the proposed method under two missing mechanisms. The values of parameters $r$ and $\gamma$ were varied within the ranges {1,2,3,4,5} and {0.1,1,10,20,100}, respectively. In this experiment, the missing rate of the data was set to 10\%. Figs.4 and 5 present the performance of the proposed algorithm with different parameters under completely random and non-random missing mechanisms for the two datasets, respectively.

From the figures, it can be observed that, for the Wine dataset with completely random missing, the proposed algorithm achieved the best performance when $\gamma$ was set to 20, and $r$ was set to 3 or 5. Conversely, for the Wine dataset with non-random missing, the optimal performance was obtained when $r$ was set to 5, and $\gamma$ was set to 10 or 20. In the case of the heart dataset with MCAR missing, the algorithm achieved the highest result when $r$ was set to 5, and the algorithm's performance was not significantly sensitive to changes in $\gamma$. For the heart dataset with MNAR missing, the algorithm achieved the highest result when $r$ was set to 5. Overall, the algorithm demonstrated relatively low sensitivity to variations in the two parameters.

\begin{figure*}[!]
  \centering
  \includegraphics[width=0.75\linewidth]{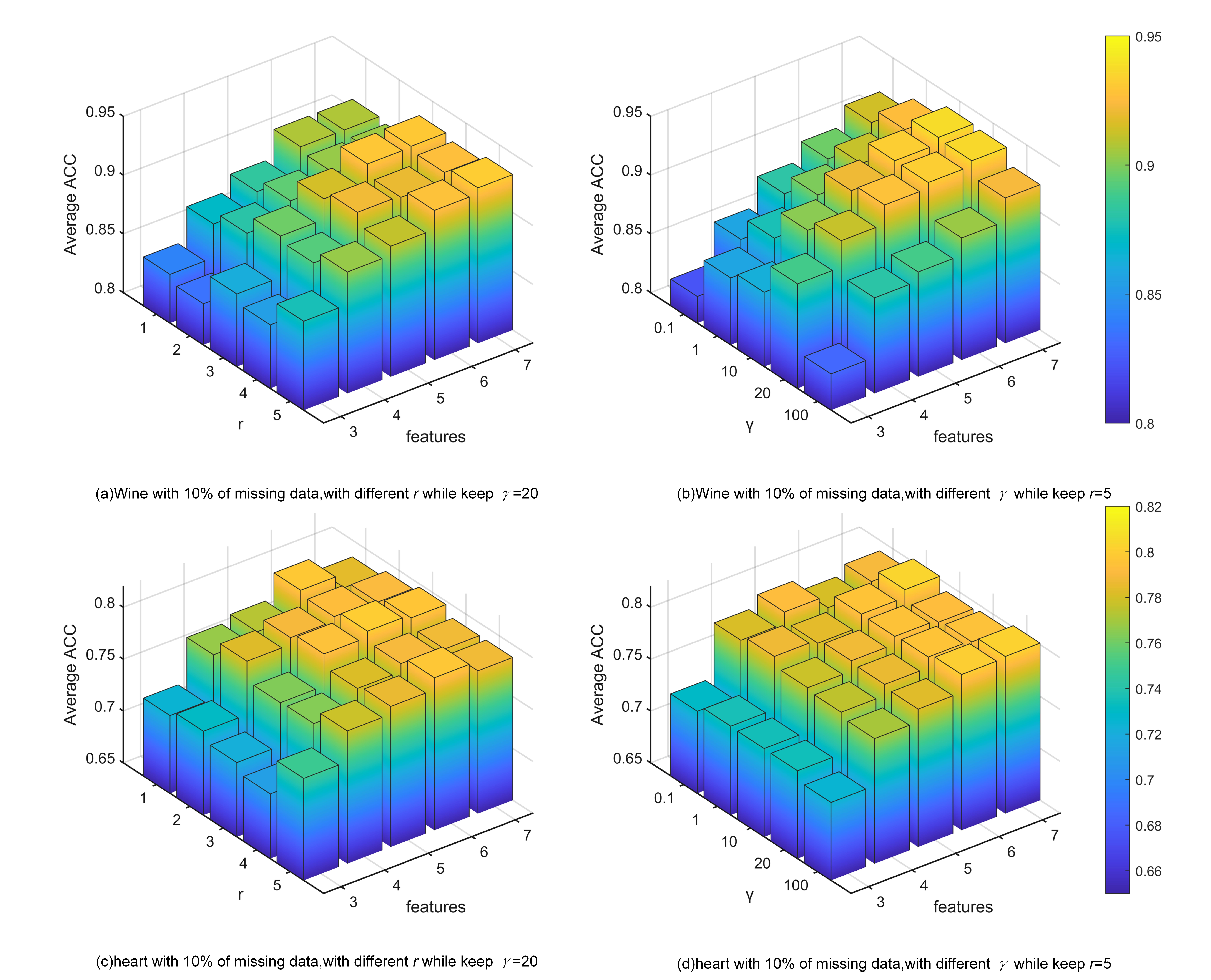}
  \caption{Accuracy of the proposed method on the Wine and Heart datasets under MCAR for different parameter settings.}\centering
\end{figure*}

\begin{figure*}[!]
  \centering
  \includegraphics[width=0.75\linewidth]{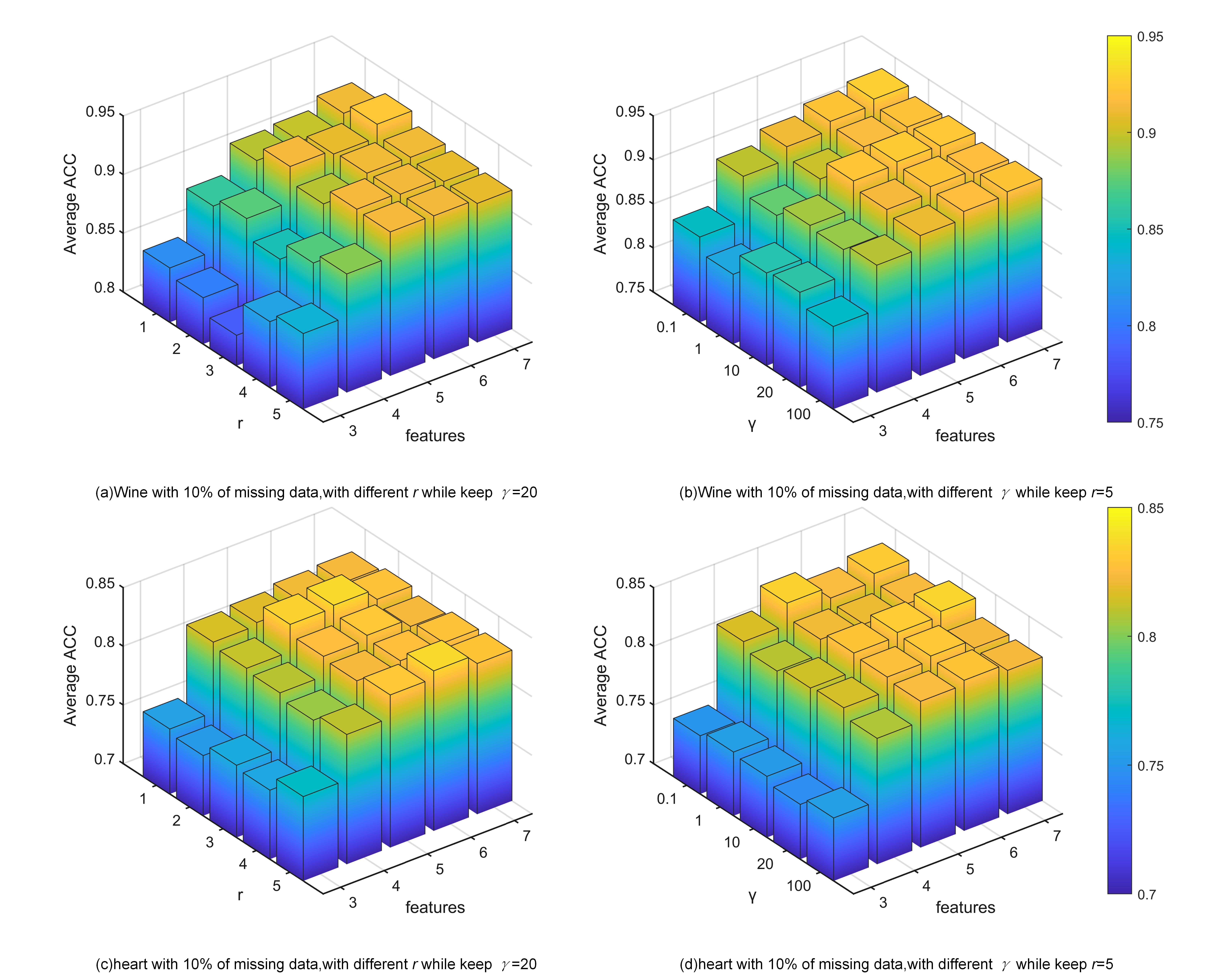}
  \caption{Accuracy of the proposed method on the Wine and Heart datasets under MNAR for different parameter settings.}\centering
\end{figure*}

\subsubsection{The Convergence of EWMC}
In this section, we investigated the convergence of the EWMC method on four datasets—Wine,Spect,heart,and Thoracic Surgery—with artificially generated 5\% of missing data. It is important to note that, for a given dataset's training and testing split, the feature importance vectors of EWMC method on the training set are different at each iteration. The objective function also varies at each iteration. Therefore, demonstrating the convergence of the EWMC method on the training dataset is challenging. To address this, we applied the feature importance vector obtained from the training dataset to the testing set and directly assessed the convergence of the EWMC method on the testing set. Fig.6 illustrates the convergence of the EWMC method on the four datasets. From the figure, it is evident that the EWMC method converges rapidly on all four datasets.

\begin{figure*}[t]
  \centering
  \includegraphics[width=\linewidth]{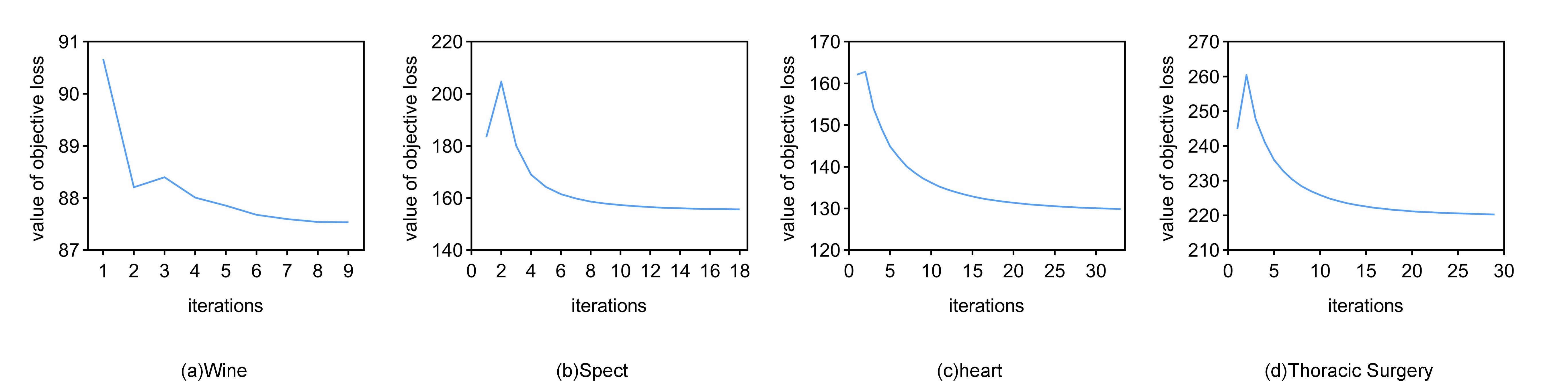}
  \caption{Convergence curves of EWMC on the four datasets with 5\% MCAR data.}\centering
\end{figure*}

\section{Conclusion}
This paper proposes a feature selection framework for incomplete data. The framework consists of two iterative stages: the matrix imputation stage and the weight learning stage,through the iteration of these two stages, feature selection is performed on the incomplete dataset. The matrix imputation stage aims to complete the data based on the current feature importance using the EWMC method. In the weight learning stage, we employ an improved ReliefF-based feature selection algorithm, $\mu$-reliefA, as the learning algorithm for this stage. The weight vector ${\bf{w}}$ obtained in the weight learning stage serves as input for the subsequent iteration of the matrix imputation stage. Results on artificially generated and real-world incomplete datasets indicate that EWMC+$\mu$-reliefA outperforms other comparative methods, demonstrating higher classification accuracy.

In this paper, we apply the proposed framework to incomplete single-label datasets. Extending this work further, one could explore the application of the algorithm to incomplete multi-label datasets by selecting multi-label feature selection algorithms during the weight learning stage. This serves as a potential avenue for future expansion of this research.

\section*{Author Contributions}
Cong Guo:Writing-original draft,Conceptualization,Validation,Methodology;Wei Yang:Writing-review and editing,Supervision.
\section*{Declarations}
{Conflicts of Interest.}
The authors declare that they have no known competing financial interests or personal relationships that could have appeared to influence the work reported in this paper.
\section*{Availability of data and materials.}
The datasets used in this study was obtained from a publicly available repository and they are available in the following website:
http://archive.ics.uci.edu/ml/index.php

%Bibliography

\end{document}